\RequirePackage{tikz} 

\documentclass[pdflatex, sn-mathphys]{sn-jnl}

\jyear{2022}

\usepackage{listings}
\usepackage{amsmath}
\usepackage[utf8]{inputenc}
\usepackage{url}
\usepackage{hyperref}
\usepackage{paralist}

\usepackage{tikz}
\usetikzlibrary{backgrounds}
\usetikzlibrary{calc}

\usepackage{paper}
\usepackage{code}

\usepackage[disable]{todonotes}
\setuptodonotes{fancyline}
\makeatletter
\define@key{todonotes}{A}[]{%
    \setkeys{todonotes}{author=Albert,color=green!30}}%
\define@key{todonotes}{G}[]{%
    \setkeys{todonotes}{author=Georg,color=red!30}}%
\define@key{todonotes}{T}[]{%
    \setkeys{todonotes}{author=Tobias,color=blue!30}}%
\makeatother

\usepackage[final]{showlabels}

\begin{document}

\title{Rule Learning by Modularity}

\author[1,2]{\fnm{Albert} \sur{Nössig} \email{albert.noessig@uibk.ac.at}}

\author[2]{\fnm{Tobias} \sur{Hell} \email{tobias.hell@uibk.ac.at}}
\equalcont{These authors contributed equally to this work.}

\author[1]{\fnm{Georg} \sur{Moser} \email{georg.moser@uibk.ac.at}}
\equalcont{These authors contributed equally to this work.}

\affil[1]{\orgdiv{Department of Computer Science}, \orgname{University of Innsbruck}, \orgaddress{\street{Technikerstrasse 21a}, \city{Innsbruck}, \postcode{6020}, \state{Tyrol}, \country{Austria}}}

\affil[2]{\orgdiv{Department of Mathematics}, \orgname{University of Innsbruck}, \orgaddress{\street{Technikerstrasse 13}, \city{Innsbruck}, \postcode{6020}, \state{Tyrol}, \country{Austria}}}

\abstract{In this paper, we present a modular methodology that combines state-of-the-art methods in (stochastic) machine learning with traditional methods in rule learning to provide efficient and scalable algorithms for the classification of vast data sets, while remaining explainable. Apart from evaluating our approach
on the common large scale data sets \emph{MNIST}, \emph{Fashion-MNIST} and \emph{IMDB}, 
we present novel results on explainable classifications of dental bills. The latter case study
stems from an industrial collaboration with  \emph{Allianz Private Krankenversicherungs-Aktiengesellschaft} which is an insurance company offering diverse services in Germany.}


\maketitle
\section{Introduction}
\label{Introduction}

There has been an immense progress in the field of machine learning in the last years. In particular, deep learning has led to the development of very complex models which yield highly accurate results but lack explainability and interpretability. To overcome this defect, the field of \emph{Explainable Artificial Intelligence} (\emph{XAI} for short, see for example~\citep{xai, xai2, Storey:2022}) has recently significantly increased momentum.
Indeed, in many application areas of machine learning, like automotive, medicine, health and insurance industries, etc.,\ the need for security and transparency of the applied methods is not only preferred but increasingly often of utmost importance. Arguably, the lack of progress in self-driving cars can be attributed to unintuitive decisions in autonomous vehicles in edge cases, that is,
situations that have not been forseen in training. In such situations safety of the behaviour of autonomous vehicles cannot currently be guarenteed, cf.~\cite{Kirkpatrick:2022}.

A textbook example of XAI is \emph{Inductive Logic Programming} (\emph{ILP})~\citep{ilp-comp}, a  subfield  of  machine  learning which aims to learn (higher-order) logic programs from input/output examples. ILP has been studied extensively over the past decades and the area encompasses a
significant number of application areas. A major advantage of ILP in this context is its transparency which makes it clearly comprehensible. While ILP is often conceived as a subfield of inductive programming, our
interest stems from the fact that logic programs are (by definition) nothing else but sets of clauses, that is, rules. Thus, a strongly related topic is \emph{Rule Induction}~\citep{F12}, where the focus is on learning simple if-then-else rules rather than finding complex programs as in modern ILP systems.
In the sequel, we refer to ILP and rule induction simply as \emph{rule learning} as our main interest
is in explainable classification for (large-scale) benchmarks, where explainability is obtained
by learning easily comprehensible rules, cf.~Section~\ref{sec:insel}.

In this paper, we present a modular methodology that combines state-of-the-art methods in (stochastic) machine learning with traditional methods in rule learning to provide efficient and scalable algorithms for the classification of vast data sets, while remaining explainable. Apart from evaluating our approach
on common large scale data sets like
\begin{inparaenum}[(i)]
  \item \emph{MNIST digits}~\citep{lecun2010mnist};
  \item \emph{Fashion-MNIST}~\citep{fmnist}; and
  \item \emph{IMDB}~\citep{imdb},
\end{inparaenum}
we also present novel results on explainable classifications of dental bills. The latter case study
stems from an industrial collaboration with  \emph{Allianz Private Krankenversicherungs-Aktiengesellschaft (APKV)} which is an insurance company offering diverse services in Germany.

More precisely, we make the following contributions:
\begin{inparaenum}[1)]
  \item We introduce a novel \emph{modular methodology}, splitting the classification of large scale
  data sets into three independent phases:
  \begin{inparaenum}[(i)]
    \item \emph{representation learning}; 
    \item \emph{input selection}; and finally
    \item \emph{rule learning}.
  \end{inparaenum}
  (See Sections~\ref{sec:meth} and~\ref{sec:insel} for further details.)
\item We provide ample experimental evidence that our modular methodology provides significant
  improvements on the efficiency of the learners over the state-of-the-art for the standard benchmarks
  (see Section~\ref{sec:eval}).
\item Further, we show that our approach yields accuracies close to stochastical, non-explainable classification methods for the case study. We emphasise that our classification yields comprehensible rules, of direct interest to our industrial collaboration partner
  (see Section~\ref{sec:caseStudy}).
\end{inparaenum}

Arguably this work also refutes a seemingly common missunderstanding that ILP methods are not
suitable to handle vast data sets. For example, Mitra and Baral~\citep{mitra18} argue that for a long time it has not been possible to apply ILP methods on benchmark data sets like the \emph{MNIST digits}~\citep{lecun2010mnist}. We not only improve upon their results for MNIST significantly, but also take
into account more sophisticated data sets like Fashion-MNIST and a sentiment analysis
for IMDB. Furthermore, we apply our approach on a real case study of industrial interest.

\paragraph*{Overview.}

In Section~\ref{sec:meth}, we introduce the aforementioned modular methodology, while we
instantiate the methodology concretly in Section~\ref{sec:insel} for the data sets studied.
Sections~\ref{sec:eval} and~\ref{sec:caseStudy} provide ample evidence of the advantages of our approach
and present the case study mentioned. In Section~\ref{sec:alternative}, we give a review of some additional machine learning tools which might be interesting for future work as well as modern ILP-methods which struggle to handle the data considered in our experiments. Finally, in Section~\ref{Related} and Section~\ref{sec:future}, we discuss related work and ideas for future work. 
To keep the presentation succinct, we have delegated the description and comparison
of the employed rule learning techniques, namely~\foil, \ripper\ and~\irep, to the Appendix, see
Section~\ref{Techniques}.

\section{Methodology}
\label{sec:meth}

\tikzstyle{txtlabel} = []
\tikzstyle{method} = [shape=rectangle, rounded corners, draw, fill=white, minimum width=2.2cm, minimum height=1.2cm]
\begin{figure}[t]
\centering
\begin{minipage}{\linewidth}
\begin{center}
  \begin{tikzpicture}[scale=.8]
  \begin{scope}
    \node[method] (represent) at (0,0) {
      \begin{minipage}{2.5cm}
        \textbf{\small Representation Learning}
      \end{minipage}};
    
    \ibox{(0,-1.5)} {2.2}{1}{\strut{}...};
    \ibox{(0.1,-1.9)}  {2.2}{1}{\tiny Neural Networks};
    \ibox{(0.2,-2.3)}  {2.2}{1}{\tiny UMAP};
  \end{scope}
  \begin{scope}
  \node[method] (clustering) at (4cm,0) {
    \begin{minipage}{2.5cm}
      \textbf{\small Input Selection}
    \end{minipage}
  };
    \ibox{(4,-1.5)} {2.2}{1}{\strut{}...};
    \ibox{(4.1,-1.9)}  {2.2}{1}{\tiny $k$-means};
    \ibox{(4.2,-2.3)}  {2.2}{1}{\tiny DBSCAN};  
  \end{scope}
  \begin{scope}
  \node[method] (learner) at (8cm,0) {
    \begin{minipage}{2.5cm}
      \textbf{\small Rule Learner}
    \end{minipage}
  };
   \ibox{(8,-1.5)} {2.2}{1}{\strut{}...};
    \ibox{(8.1,-1.9)}  {2.2}{1}{\tiny \foil};
    \ibox{(8.2,-2.3)}  {2.2}{1}{\tiny \ripper};
  \end{scope}
  \node[minimum width=1.3cm,anchor=west,align=left] (answer) at ($(learner.east)+(4mm,0)$) {    
\begin{lstlisting}[style=prolog,basicstyle=\color{darkviolet}\scriptsize\ttfamily,frame=none]
target(V) :-
  black_1(V),
  black_N(V).
\end{lstlisting}
  };
  \begin{scope}[on background layer]
    \draw[->] (represent) -- (clustering);
    \draw[->] (clustering) -- (learner);
    \draw[->] (learner) -- (answer);
  \end{scope}
\end{tikzpicture}
\end{center}
\end{minipage}
\vspace{1ex}
\caption{Modular Approach to Rule Learning.}
\label{fig:approach}
\end{figure}
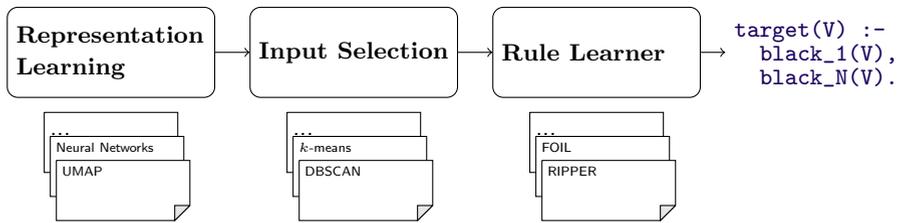

Up to now, ILP methods have seldomly been combined with state-of-the-art methods in machine learning, with the possible exception of earlier
work by Evans et al.~\cite{dILP} and Payani~et al.~\cite{dNL}. In~\cite{dILP,dNL} differential deep neural logic networks have been
used to learn explanatory rules even in the presence of noisy data. Similar to the modular approach to rule learning that
we introduce in this paper, Eldbib~\cite{Eldbib:2016} introduced a so-called \emph{Data Grouping Method} which makes it possible to arrange the given examples in a beneficial order. However, this particular approach has been developed for a certain family of rule learners which only considers one example at a time and, therefore, especially benefits from a meaningful order of input examples. 

In the majority of the literature, however, ILP methods have not been employed to standard benchmarks, like \emph{MNIST},
\emph{Fashion-MNIST}~\citep{fmnist}, \emph{IMDB}~\citep{imdb}, etc. that are routinely employed in unexplainable state-of-the-art methods.
In particular, modern rule induction and ILP methods are mostly not able to compete with, for instance, neural networks. 
Foremostly, this poses a technical problem but more importantly this is a conceptual challenge.

We intend to unlock the potential of ILP and Rule Induction by combining such rule learners with additional machine learning tools in order to make them applicable to a larger variety of problems. More
precisely, we aim to improve the performance of rule learning methods to ease their application to large data sets, especially in the context of explainable classification regarding our case study (cf. Section~\ref{sec:caseStudy}).
Figure~\ref{fig:approach} depicts our modular approach to this extent. The methodology is comprised of three independent phases
  \begin{inparaenum}[(i)]
    \item \emph{representation learning}; 
    \item \emph{input selection}; and finally
    \item \emph{rule learning}.
    \end{inparaenum}    

We first apply standard, state-of-the-art ML-tools like neural networks or dimensionality reduction methods such as, for instance, UMAP~\cite{umap} for \emph{representation learning}.
The resulting features represent the original data in a more compact form, eg. the $28 \times 28$ images of the MNIST data set can be reduced to their first 50 principal components as considered in the experiments discussed in Section~\ref{sec:eval}. 
This more compact form of the input data is advantagous for clustering (eg. \emph{k-means}, \emph{DBSCAN}~\cite{dbscan}) applied subsequently in order to conduct a meaningful \emph{input selection} which results in increased speed and thus scalability of the approach (see Section~\ref{sec:insel} for further details.)
The selection of beneficial input data for the applied method supports its learning process. With the example of the MNIST digits it is, for instance, easier to learn rules on different forms of zeros independently than on the whole set at once as illustrated in Figure~\ref{fig:mnist_rules} and explained in Section~\ref{sec:insel} in further detail. 
The \emph{rule learning} method such as, for instance, \foil~\citep{DBLP:journals/ml/Quinlan90} or \ripper~\citep{COHEN1995115} is eventually able to find comprehensible rules that can be used for classification, cf.~Figure~\ref{fig:approach} and Figure~\ref{fig:mnist_rules}, respectively. Considering a MNIST digit, for instance, the learner yields rules regarding the pixel intensity in order to predict the correct label.

Note that the learner uses the data in its original form, instead of the features extracted in the first step
which are only used to improve the input selection. This is crucial because otherwise we lose explainability.
Suppose, for instance, we would attempt to learn rules on features extracted
in the representation learning phase. Say we employ convolutional neural networks (CNNs) to
the analysis of, for example, the MNIST data set and subsequently learn rules from the obtained features in the penultimate layer. Naturally, the accuracy
and precision of the obtained rules would be outstanding and near to the original classification by the CNN. However, explainability has been lost: The learned rules only consider the incomprehensible features used for the black-box classification within the neural network instead of the pixel intensity.  

Summing up: To conduct the selection of input samples for rule learning, clustering has to be applied on the compact but incomprehensible features obtained from representation learning. The rule learning itself is then performed with the original, \emph{comprehensible} input variables. In this way, rule learning leads to comprehensible rules by implicitly using complex information encoded in the sample selection.

With the above-mentioned procedure, we are able to apply rule learning methods in reasonable time on extensive data sets and our approach makes it possible to reduce computation time significantly, eg. we obtain a speedup by as much as 6 on the MNIST digits (see Section~\ref{sec:eval} for further details). 
On the other hand, the conducted experiments have also shown that the achieved time savings are highly dependent on the applied learner and some of the modern ILP-tools are not applicable at all in this context. We will comment on alternative scenarios in Section~\ref{sec:alternative}.

\section{Clustering  \& Rule Learning}
\label{sec:insel}

This section describes the process of input selection in further detail and presents the advantages of clustering in this context with the example of the MNIST handwritten digits data set.
Further, we define \emph{rule learning} more precisely and give examples of common methods in this context.

\begin{figure}[t]
	\begin{center}
	\setlength{\fboxsep}{0pt}
	\fbox{
	\includegraphics[trim=240 90 200 10,clip,width = 0.9\textwidth]{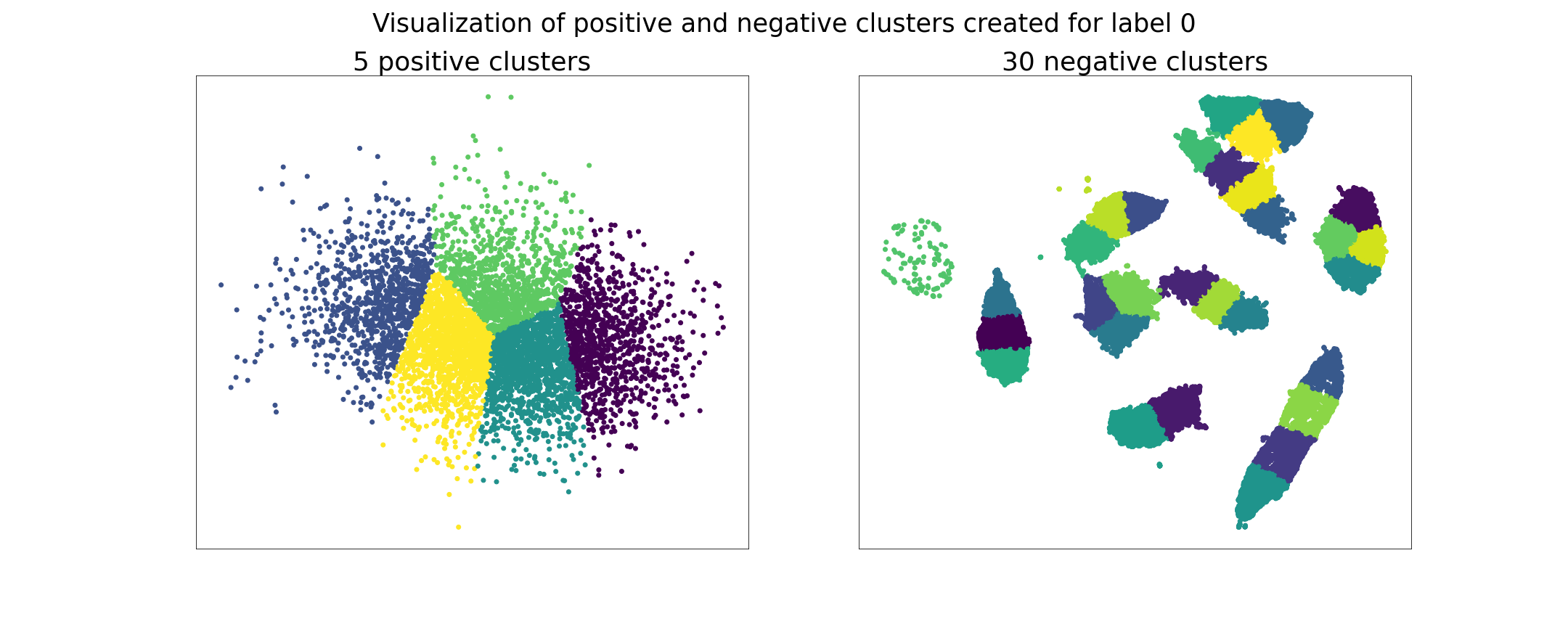}
	}
	\end{center}
	\caption{Visualisation of the first and second principal components of the positive and negative clusters obtained with the example of zeros in the MNIST data set.}
        \label{fig:clusters}
\end{figure}

\subsection{Input Selection} 

\begin{figure}[t]
	\begin{center}
	\setlength{\fboxsep}{0pt}
	\fbox{
	\includegraphics[trim=250 90 210 10,clip,width = 0.85\textwidth]{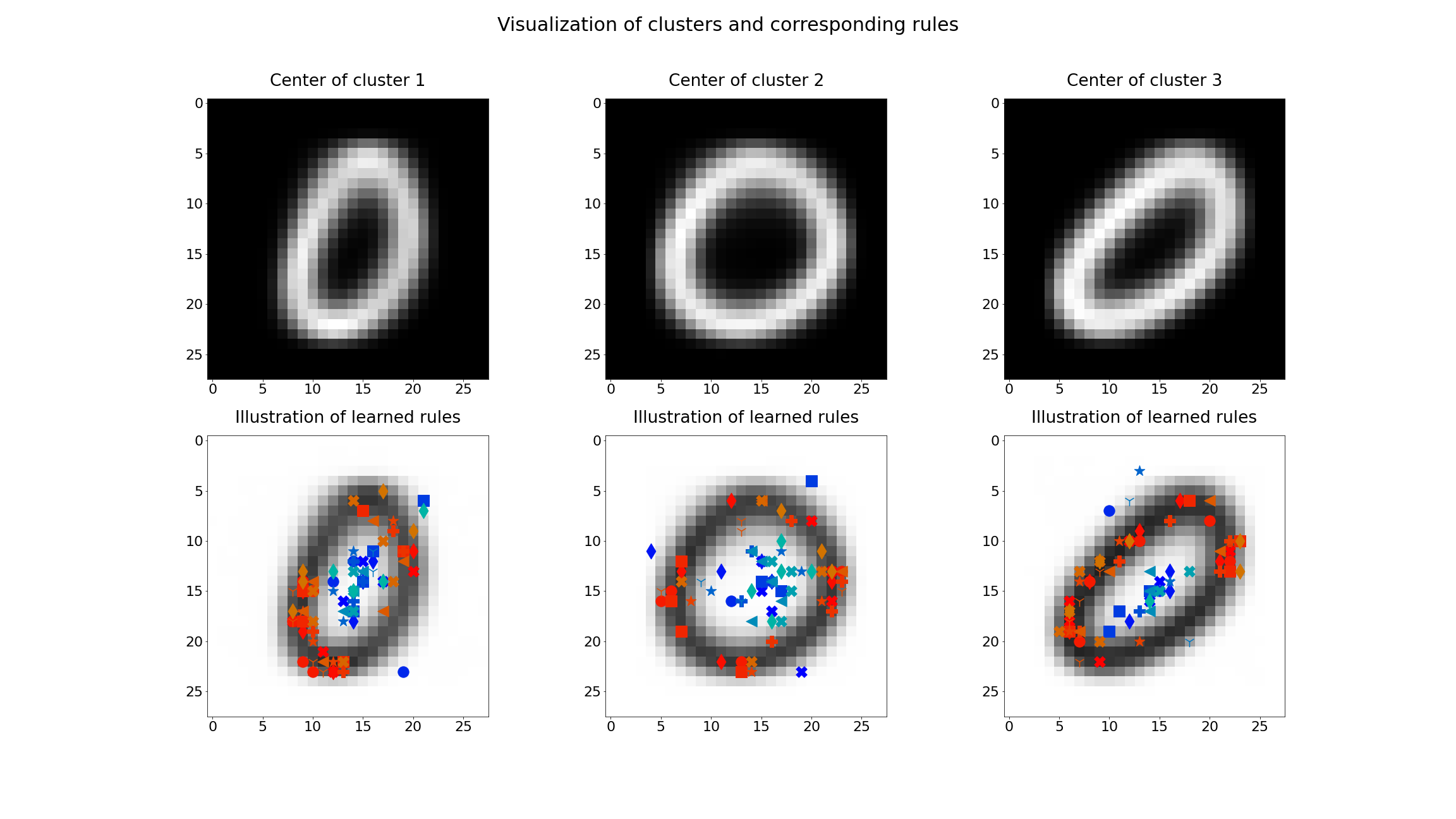}
	}
	\end{center}
	\caption{Illustration of rules learned in the context of classifying whether a given digit is equal to zero or not, where the positive examples have been diveded into 3 clusters.}
        \label{fig:mnist_rules}
\end{figure}

As already mentioned, the crucial part of our approach is the selection of beneficial input data. For this purpose, we divide the original input data into several clusters in order to make it easier for the algorithm to find the rules. Basically, we proceed as follows. 
We consider the positive and negative examples separately and apply clustering on both because we want the positive examples to be as homogeneous as possible and the negative ones as heterogeneous as possible. Figure~\ref{fig:clusters} depicts the results of clustering applied on the MNIST data set in the context of predicting whether a given digit is a zero. 
The resulting positive clusters are shown on the left side. By definition, they contain similar positive examples where it is easier to find rules describing them compared to learning rules on the whole positive examples at once, cf.~Figure~\ref{fig:mnist_rules}.
On the right-hand side, the resulting negative clusters are shown. They are used to reduce the number of negative examples given to the learner. Depending on the size of the considered data set, we arbitrarily extract a certain percentage of each cluster (by default 25\%) and join the chosen \emph{negative representatives} together. By doing so, we obtain a set of meaningful, distinctive negative examples while reducing the problem size and consequently the complexity of the rule space by a multiple.
As a next step, we append the chosen negative representatives from above to each of the positive clusters. In this way, we obtain several independent sets of examples serving as input for a subsequently applied rule learner.

\paragraph*{Advantages of Clustering}

As mentioned above, splitting the positive examples into clusters of similar examples eases the process of rule learning. In particular, the example of the MNIST digits illustrates the advantages of clustering.
Figure~\ref{fig:mnist_rules} shows that we can split the examples representing a zero into groups of different ways of writing it. The first cluster consists of images of \emph{straight} zeros, the second cluster represents \emph{round} zeros and the third \emph{slanted} zeros. By focusing on each cluster individually, we enable the method to learn the rules more efficiently as shown in Section~\ref{sec:eval} and illustrated in the figure, where the markers in different red shades correspond to an expected positive pixel intensity and the blue shades to zero pixel intensity. Each marker style illustrates a single rule and it is noticable that the learned rules on each cluster exactly match with the form of the corresponding zeros. For instance, the rules corresponding to the cluster on the left all require chosen pixels in an oval form to have positive pixel intensity with a quite elongated field of white pixels in the middle. These rules describe the form of the straight zeros very well. On the contrary, the rules corresponding to the cluster in the middle require pixels in a broad and round form to have positive pixel intensity with white pixels evenly spread in circular form within them matching with the round zeros contained in this cluster.

For the examples in Section~\ref{sec:eval}, k-means clustering has shown to perform best but other methods such as DBSCAN \citep{dbscan} might be advantageous in different problem settings and are consequently also applicable within our modular approach.

\subsection{Rule Learning}
\label{RuleLearning}

\emph{Inductive Logic Programming} investigates  the  construction  of  first- or higher-order  logic  programs.
In the context of this paper, it suffices to conceive the learnt hypothesis as first-order Prolog clauses as depicted below.
\begin{lstlisting}[style=Prolog]
  H :- L1, ..., Lm
\end{lstlisting}
Here, the \emph{head} $H$ is an atom and the \emph{body} $L_1, \dots, L_m$ consists of literals, that is atoms or negated atoms. An extensive overview on ILP is given by Cropper et al.~\citep{ilp-comp}. A selection of commonly used ILP methods, together with some crucial properties, are shown in Table~\ref{tab:ilp-comp}. 

One of the first ILP systems was \foil\ by Quinlan~\citep{DBLP:journals/ml/Quinlan90}. Regarding ILP, we investigate mostly this algorithm due to its simplicity and the fact that our \emph{Python} reimplementation (cf.~Section~\ref{Techniques}) is able to handle large data sets straight away. Moreover, it seems to be the only ILP method which is effectively working on our large benchmark data sets.
The applicability of the remaining methods in Table~\ref{tab:ilp-comp}  to the considered data sets is investigated in Section~\ref{sec:alternative}. 
Note that we do not make use of the full potential of ILP. Instead of learning complex programs we aim to learn simple if-then-else rules like Rule Induction methods tend to. For instance, we predict a given image to be equal to zero if specific pixels are black as shown above and explained in more detail in Section~\ref{sec:eval}. 

A strongly related topic is \emph{Rule Induction} focusing on learning simple if-then-else rules such as the following example by Fürnkranz et al. \cite{F12}. A data set containing the variables \emph{Education, Marital Status, Sex, Has Children} and \emph{Car} is given and the aim is to learn rules concerning the first four variables to predict the type of car. For instance, the following rule is learned:

\begin{lstlisting}[style=Prolog]
		IF	Sex = Male
		  AND	Has Children = Yes
		THEN	Car = Family.

\end{lstlisting}

Many techniques used to learn sets of decision rules have been adapted from decision tree learning and have been enhanced especially to better handle noisy data. In this context, we particularly consider the state-of-the-art method \ripper\ \citep{COHEN1995115} and its predecessor \irep\ \citep{Frnkranz1994IncrementalRE} in our experiments (cf. Section~\ref{sec:eval} and Section~\ref{Techniques}). More rule induction methods are outlined by Scala et al. \citep{Scala2019KnowledgeGW} who applied them in the context of cancer research.

\renewcommand{\arraystretch}{1.1}
\begin{table}[t]
\centering
\begin{tabular}{@{}|@{\,}l@{\,}|@{\,}c@{\,}|@{\,}c@{\,}|@{\,}c@{\,}|@{\,}c@{\,}|@{\,}c@{\,}|@{\,}c@{\,}|@{\,}c@{\,}|@{\,}c@{\,}|@{\,}c@{\,}|@{\,}c@{\,}|@{}}
 \hline
  \textbf{System} & \foil & \lime & \progol & \alephsystem & \ilasp & \metagol & \dilp & \dnl & \fastlas & \popper
  \\[.5mm] \hline
  \textbf{Noise} & \YES & \YES & \YES & \YES & \YES & \NO & \YES & \YES & \YES & \YES
  \\[.5mm] \hline
  \textbf{Optimality} & \NO & \NO & \NO & \NO & \YES & \YES & \YES & \YES & \YES & \YES
  \\[.5mm] \hline
  \textbf{Domains} & \INF & \INF & \INF & \INF & \NOINF & \INF & \NOINF & \NOINF & \NOINF & \INF
  \\[.5mm] \hline
  \textbf{Hypotheses} & LP & LP & LP & LP & ASP & LP & LP & LP & ASP & LP
  \\[.5mm] \hline
\end{tabular}
\smallskip
\caption{Comparison of popular ILP systems. LP simply refers to (a subset) of Prolog rules, while ASP abbreviates answer-set programming.}
\label{tab:ilp-comp}
\end{table}

Independent of the chosen learner, we apply the method on each set of input examples given by the input selection individually. This way of proceeding is easily parallelisable and results in significant time savings as shown in Section~\ref{sec:eval}.

\section{Evaluation}
\label{sec:eval}

In the previous sections, we have introduced our methodology, which we evaluate in the following by applying 
it on some well-known benchmark data sets as well as a practical example taken from insurance industries. At first, we consider the above-mentioned digits and explain the progress of our experiments in further detail.

\paragraph{MNIST~\citep{lecun2010mnist}.}

This data set consists of $28 \times 28$ greyscale images of digits (cf. Figure \ref{fig:mnist_rules}) which are split into 60 thousand training examples and 10 thousand test examples, where all digits are evenly represented.
In related work, Mitra et al.~\citep{mitra18} have investigated the use of answer set programming (ASP) for the
classification of MNIST digits. As reported in~\citep{mitra18} they have not been able to process the MNIST digits in their original form.
Thus, we have been curious in what extent we have to adapt the original data set in order to make it suited for our approach.

\begin{figure}[t]
	\begin{center}
	\fbox{
	\includegraphics[trim=160 30 110 20,clip,width = 0.9\textwidth]{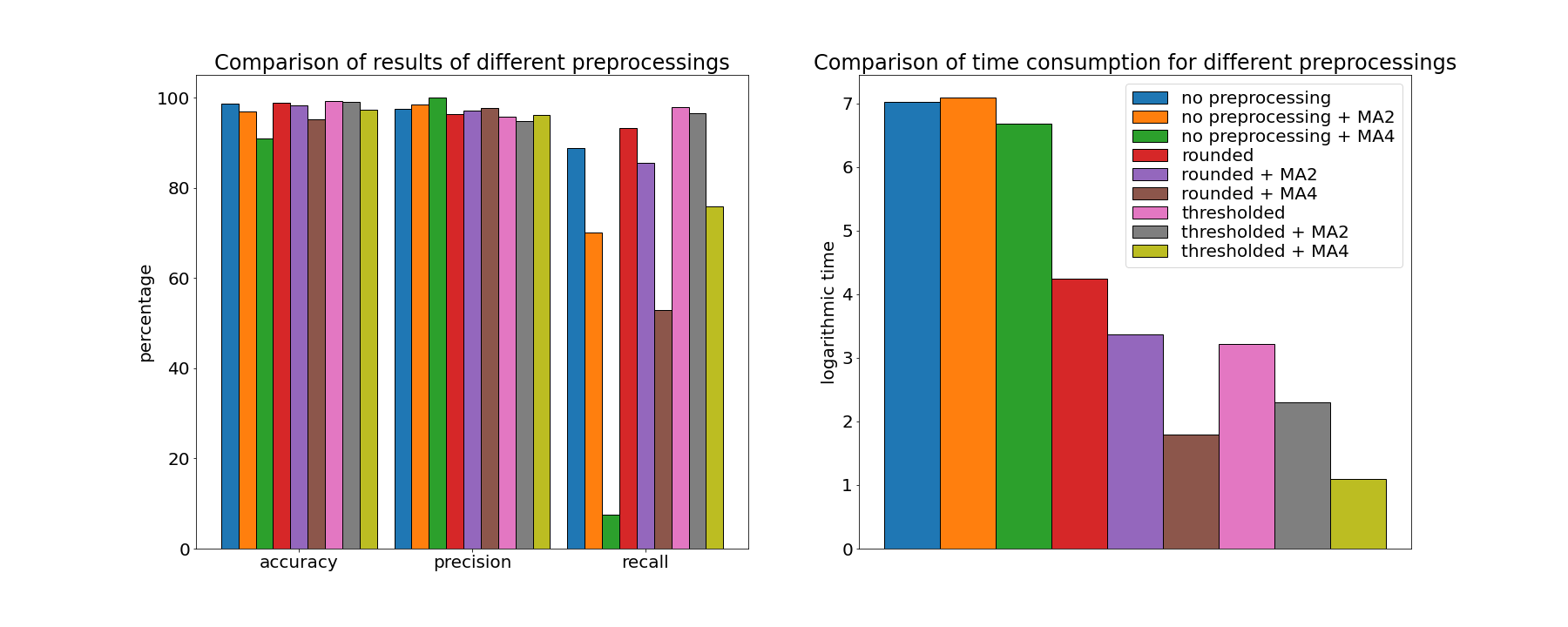}
	}
	\end{center}
	\caption{Comparison of results corresponding to different types of preprocessing on the MNIST digits data set.}
        \label{fig:preprocessing}
\end{figure}

Figure~\ref{fig:preprocessing} illustrates the results of our considered steps of preprocessing. We start with the original input data transformed into vectors of length 784 by way of rounding the pixel intensity to encoding the pixels in \emph{black} and \emph{white} by applying thresholding. Moreover, we combine the above preprocessing steps with the calculation of a moving average (denoted with \emph{MA2} for filter size $2 \times 2$ or \emph{MA4} for filter size $4 \times 4$) inspired by Mitra and Baral, who built submatrices of size $7 \times 7$ by summing up the values in each square of size 4 in their work \citep{mitra18}. We come to the conclusion that it is not only possible to apply the \foil\ algorithm on the MNIST digits in their original size of $28 \times 28$, but it also yields better results than the ones obtained by the application of submatrices. 

Due to the fact that the rules learned on the full-size images thresholded into black and white pixels are the most expressive in our opinion and the corresponding results are consistently among the best ones, we stick with this kind of preprocessing and conduct the further experiments on the corresponding binary vectors of length 784. The mentioned expressiveness and comprehensibility of the rules learned on these vectors is illustrated in Figure~\ref{fig:mnist_rules} which also indicates the straightforward application of the rules. Each rule can be used like a filter. We put it over the given image of a digit to be predicted and according to whether the respective pixels are black or white, we make our decision. On the other hand, we can also express the rules as shown with the following practical example, where \emph{target} for instance means that the given image \emph{V} depicts a zero.

\begin{lstlisting}[style=prolog]
target(V) :- black_215(V), white_325(V), white_382(V), black_386(V), 
						 black_443(V), white_462(V), white_466(V), white_719(V).
\end{lstlisting}

\begin{figure}[t]
	\begin{center}
	\fbox{
	\includegraphics[trim=170 35 150 25,clip,width = 0.9\textwidth]{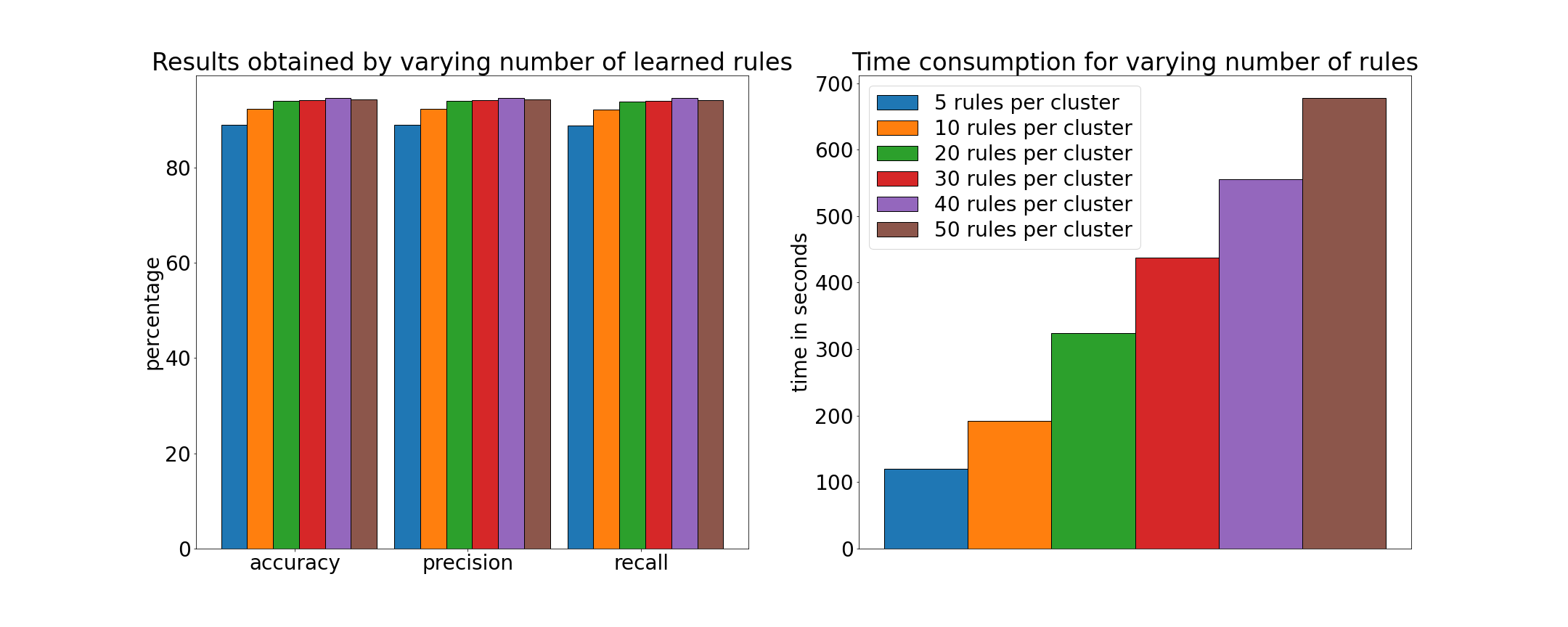}
	}
	\end{center}
	\caption{Comparison of results corresponding to varying number of learned rules on 3 positive clusters.}
        \label{fig:nrOfRules3}
\end{figure} 

\begin{figure}[t]
	\begin{center}
	\fbox{
	\includegraphics[trim=170 35 150 25,clip,width = 0.9\textwidth]{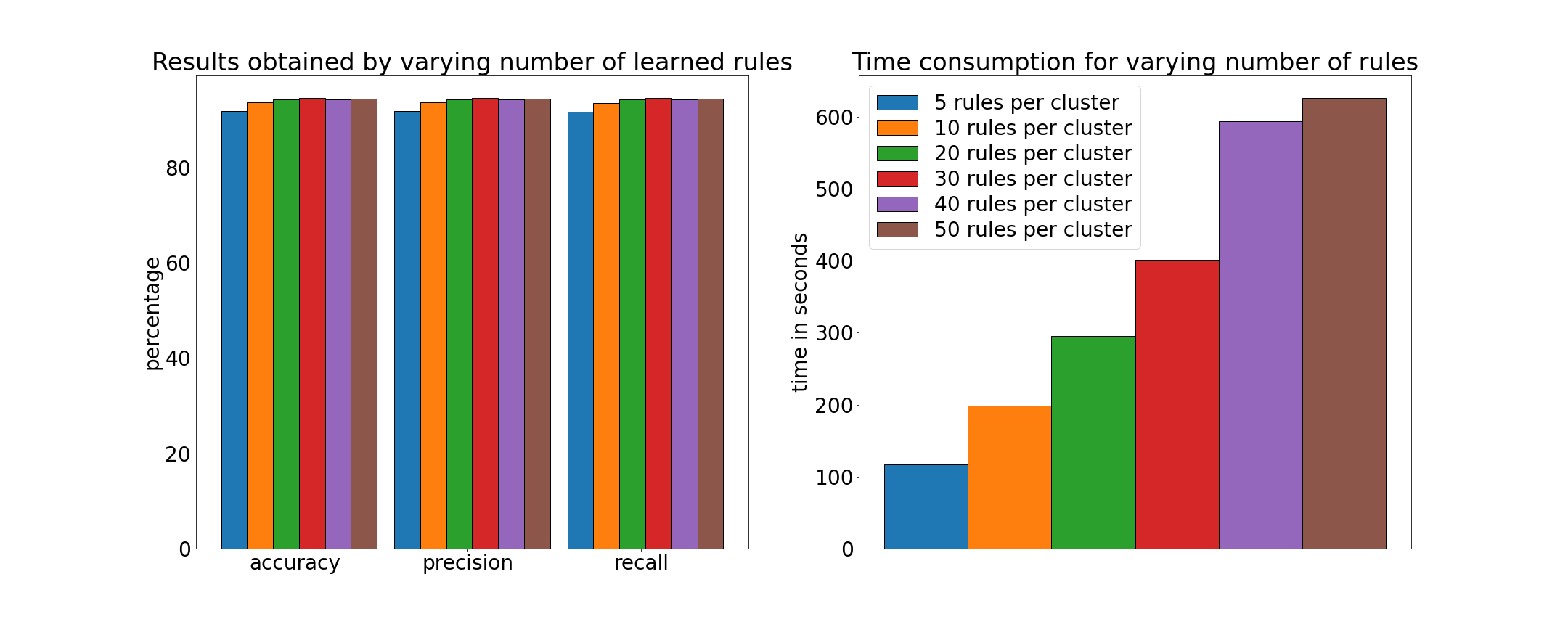}
	}
	\end{center}
	\caption{Comparison of results corresponding to varying number of learned rules on 5 positive clusters.}
        \label{fig:nrOfRules5}
\end{figure} 

After choosing an appropriate way of preprocessing, we need to decide how many rules we want to learn on each cluster in our experiments. We examine the effect of varying number of rules on the resulting accuracy with the example of 3 and 5 clusters, respectively. The corresponding results for the prediction of a zero are shown in Figure~\ref{fig:nrOfRules3} and Figure~\ref{fig:nrOfRules5}. 

Naturally, the time consumption increases with increasing number of learned rules. On the other hand, we observe that the results do not change significantly for more than 20 rules. Therefore, we learn at most 20 rules on each cluster in order to obtain high accuracies while keeping the time consumption on a moderate level. In addition, the comprehensibility would decrease with a higher number of learned rules.

Obviously, the total number of learned rules strongly depends on the chosen number of applied positive clusters. As we figured out that about 20 rules are advantageous for describing the features of a specific cluster, we use not more than 5 positive clusters to keep the total number of rules on a moderate level. Moreover, further increasing the number of positive clusters would result in very small clusters which might lack expressiveness.

Table \ref{tab:mnist} shows the results from \cite{mitra18} compared with our results. The first row shows the percentage of correctly identified examples, ie. \emph{recall}, on each digit achieved by Mitra et al.%
\footnote{We interpret the percentage of correctly classified results presented in~\cite{mitra18} as \emph{recall} and correspondingly compare to
  our results on recall. On the other hand, Mitra et al.\ depict the corresponding percentages as ``accurracy'',
  which however in our opinion, is inconsistent with the gist of the presented result. Furthermore, our accurracy results are even more distinguished. We obtain at least 98\% for each digit.}
Analogously, the rows below show our outcomes using 5 positive clusters for each digit. Note that their approach takes 160 hours to learn the rules for each digit (except of the digit one), whereas our approach has a total time consumption between only a few minutes with \foil\ as chosen learner and under 2 hours when \ripper\ is applied as shown in further detail in Table~\ref{tab:results}. Moreover, they learn thousands of rules for each digit, whereas we restrict ourselves to learn at most 20 rules on each cluster. 

In addition to testing our approach on the MNIST digits, where unexplainable state-of-the-art methods for multi-class classification achieve an accuracy of more than 99\% \citep{DBLP:journals/corr/abs-2008-10400}, we also consider \emph{Fashion-MNIST} and \emph{IMDB} as
benchmark data sets (see Table~\ref{tab:results} for the
results.)

\renewcommand{\arraystretch}{1.3}
\begin{table}[t]
\centering
\begin{tabular}{@{}|l|@{\,}c@{\,}|@{\,}c@{\,}|@{\,}c@{\,}|@{\,}c@{\,}|@{\,}c@{\,}|@{\,}c@{\,}|@{\,}c@{\,}|@{\,}c@{\,}|@{\,}c@{\,}|@{\,}c@{\,}|@{}}
\hline
  & 0 & 1 & 2 & 3 & 4 & 5 & 6 & 7 & 8 & 9
  \\[.5mm] \hline
  Mitra et al. & 60,91 & 95,85 & 32,95 & 49,80 & 49,59 & 42,65 & 65,03 & 63,52 & 54,50 & 69,18
  \\[.5mm] \hline
  \foil\ & 98,10 & 98,44 & 92,11 & 92,17 & 92,74 & 91,97 & 95,42 & 93,21 & 87,50 & 86,82
  \\[.5mm] \hline
  \irep\ & 98,06 & 97,87 & 92,67 & 90,85 & 90,54 & 93,53 & 95,11 & 92,27 & 90,90 & 90,05
  \\[.5mm] \hline
  \ripper\ & 98,71 & 97,92 & 95,15 & 92,99 & 94,25 & 93,04 & 97,54 & 95,10 & 92,60 &92,42
  \\[.5mm] \hline
\end{tabular}
\smallskip
\caption{Performance of our approach on handwritten digit recognition tasks compared with \cite{mitra18}.} \label{tab:mnist}
\end{table}

\paragraph{Fashion-MNIST~\citep{fmnist}.}

This data set is very similar to the MNIST digits data set. It consists of $28 \times 28$ grayscale images of pieces of clothing, where we use the same preprocessing as above, ie. we encode the pixels as \emph{black} and \emph{white}. We stick with the same maximal number of rules and learn 20 rules on each cluster. Note that unexplainable state-of-the-art methods achieve an accuracy of about 96\% on this data set \citep{DBLP:journals/corr/abs-2006-09042}.

\paragraph{IMDB Movie Reviews~\citep{imdb}.}

This data set consists of informal movie reviews from the \emph{Internet Movie Database}, where unexplainable state-of-the-art methods achieve an accuracy of around 97\% \citep{imdb_best}. We transform the underlying text data into binary vectors representing the 1000 most common words in the data set. So, we consider whether one of these words occurs in the text or not and use this information for sentiment analysis. Again, we learn at most 20 rules on each cluster.
 
Table \ref{tab:results} shows the results of our methodology applied on the above-mentioned data sets. Note that the depicted results correspond to multiclass classification. As opposed to the results illustrated in Table~\ref{tab:mnist}, we not only learn rules for binary classification stating, for instance, whether a given digit is equal to zero or not, but combine the binary classifiers trained on each possible label to a multiclass classifier. In order to do that, we construct a binary vector of length equal to the number of labels. Each entry represents the prediction of a rule set learned in advance to classify whether the corresponding label is consistent with the given input. For instance, in the case of the MNIST digits assume that an image depicting a zero is given to the multiclass classifier. The optimal output for this example is the vector
$(1,0,0,0,0,0,0,0,0,0),$
where only the first set of rules learned especially for the label \emph{zero} predicts \emph{True} and all the other rule sets corresponding to the remaining digits predict \emph{False}.

It might occur that two or more binary classifiers return \emph{True} resulting in an ambiguous multiclass classification. In this case, we additionally take the length of the learned rules, eg. the number of considered pixels in the rules shown in Figure~\ref{fig:mnist_rules}, into account and return the label corresponding to the longest satisfied rule. On the other hand, if no binary classifier returns \emph{True}, we consider the percentage of satisfied partial rules, eg. how many of the pixels corresponding to a rule have the intensity as requested by the rule, and predict the label with highest percentage.

Our approach makes it possible to reduce the time consumption for learning the rules by a multiple without negatively affecting the obtained accuracy. This is a consequence of the opportunity to parallelise the learning process on each cluster. Especially the \foil\ algorithm benefits from the additional clustering resulting in a speedup of up to a factor of 44 (cf. Table~\ref{tab:bills}), but it is also advantageous for the other methods. For instance, the \ripper\ algorithm shows a speedup factor of about 3 in all examples.

In addition, the experiments indicate that our modular approach also contributes to the improvement of the results. For instance, the obtained accuracy of the \ripper\ algorithm applied on the Fashion-MNIST data set increases by about 1,3\% while the time cosumption reduces to a third.

Furthermore, we can observe that the more complex and modern \ripper\ algorithm does not show any advantages concerning accuracy over the older and simpler \foil\ algorithm when primitive \emph{black/white rules} are learned. On the other hand, our case study on dental bills (cf.~Table~\ref{tab:bills}) demonstrates that \ripper\ is superior to \foil\ when more complex input data is considered, since it is able to learn rules of the form
$\textit{attribute} > \textit{value}$ or $\textit{value}_1 < \textit{attribute} < \textit{value}_2$ instead of simply
$\textit{attribute} = \textit{value}$.

The obtained results are a consequence of a variety of experiments examining
the underlying parameters such as the number of clusters, the applied machine learning tools or the application of clustering on the negative examples. At first we experimented with the Fashion-MNIST data set---since it is more complex than the MNIST digits---resulting in some coarse restrictions. For instance, we figured out that the advantages of negative clustering depend on the chosen learner, eg. the \foil\ algorithm benefits from getting all negative examples as input, whereas the rule induction methods profit from a reasonable selection of negative examples (in the experiments above we use 25\% of the negative examples arbitrarily taken from up to 30 clusters). Moreover, DBSCAN has shown to yield no applicable results in this context because it either produces far too many very small clusters or no meaningful clusters at all. 
Afterwards, taking these restrictions under consideration, we conduct some further experiments on each of the benchmark data sets in order to find the optimal settings for each problem and the corresponding results are shown in Table~\ref{tab:results}.%
\footnote{An extended overview of the received results is available on~\url{https://github.com/AlbertN7/ModularRuleLearning}.}

\begin{table}[t]
\centering
\begin{tabular}{@{}|@{\,}l@{\,}|@{\,}l@{\,}|@{\,}r@{\,}|@{\hspace{1mm}}c@{\hspace{1mm}}|@{\hspace{1mm}}r@{\hspace{1mm}}|@{\hspace{1mm}}r@{\hspace{1mm}}|@{\hspace{1mm}}r@{\hspace{1mm}}|@{}}
\hline
                                                                                                                                       \hfil \textbf{data} \hfill & \hfil \textbf{learner} \hfill & \hfil \textbf{time} \hfill & \textbf{speedup} & \textbf{accuracy} & \textbf{precision} & \textbf{recall}
  \\ \hline\hline
MNIST & \foil\ & 1.960 s & & 94,3\% & 94,3\% & 94,2\%\\
\hline
 & \foil\ - mod. & 309 s& 6,3 &94,3\% & 94,3\% & 94,2\%\\
\hline
 & \irep\ & 2.242 s & &87,6\%  &89,4\% & 87,5\%\\
\hline
 & \irep\ - mod. & 1.500 s & 1,5 &88,6\% & 89,1\% & 88,5\%\\
\hline
 & \ripper\ & 8.292 s & &91,7\% & 92,9\% & 91,5\% \\
\hline
 & \ripper\ - mod. & 2.770 s & 3 &91,6\% & 91,8\% & 91,4\%\\
\hline
\hline
Fashion-MNIST & \foil\ & 2.540 s & &84,6\%& 84,5\% & 84,4\%\\
\hline
 & \foil\ - mod. & 469 s & 5,4 &84,6\% & 84,6\% & 84,4\%\\
\hline
 & \irep\ & 2.970 s & &79,9\% & 81,3\% & 79,9\%\\
\hline
 & \irep\ - mod. & 1.543 s & 1,9 &79,9\% & 80,4\% & 79,9\%\\
\hline
 & \ripper\ & 11.353 s & &82,0\% & 85,1\% & 82,0\%\\
\hline
 & \ripper\ - mod. & 3.703 s & 3 &83,3\% & 83,9\% & 83,3\%\\
\hline
\hline
IMDB & \foil\ &  2.221 s & &76,1\%&  77,3\% & 76,1\%\\
\hline
& \foil\ - mod. &  218 s& 10,2 &75,0\% & 75,6\% & 75,0\%\\
\hline
& \irep\ & 300 s & &72,7\% & 73,2\% & 72,7\%\\
\hline
& \irep\ - mod. &  158 s & 1,9 &69,2\% & 70,7\% & 69,2\%\\
\hline
& \ripper\ & 914 s & &71,2\% & 75,5\% & 71,2\%\\
\hline
& \ripper\ - mod. &  356 s & 2,6 &75,2\% & 75,3\% & 75,2\%\\
\hline
\end{tabular}
\smallskip
\caption{Performance of our approach on different benchmark problems. Note that \emph{mod.} denotes the combination of clustering and rule learning. Times have been measured on a CPU using AMD Ryzen Threadripper 2950X WOF.} \label{tab:results}
\end{table}

To sum up, our experiments show that \ripper\ indeed outperforms \foil\ when complex input data is considered (cf. Section~\ref{sec:caseStudy}). Nevertheless, we believe it merits attention that the \foil\ algorithm as one of the first methods in the field of ILP is able to compete with state-of-the-art rule induction methods when simple rules are learned. In particular, the obtained time savings of our approach combined with the \foil\ algorithm are tremendous. In this context, an interesting idea for future work is the development of a dedicated ILP method combining the advantages of ILP and rule induction. 

\section{Case Study: Dental Bills}
\label{sec:caseStudy}

The \emph{Allianz Private Krankenversicherungs-Aktiengesellschaft (APKV)} is an insurance company offering diverse services in Germany. In the course of digitisation, it nowadays becomes a standard to offer the possibility to hand in scans of medical bills via, for instance, an app. Those scanned bills are then automatically processed and the goal is to reimburse the approved sum as fast as possible without a person handling the case. The benefits are lower personnel costs and to gain an edge over the competition if the client recieves the reimbursed money shortly after handing in the bill. More information about the application of artificial intelligence particularly in the insurance business is given by Lamberton et al.~\citep{dunkelverarbeitung}.

However, a variety of challenges has to be tackled before an automated decision regarding the bill can be made: The quality of the scans may vary, information can be given quite differently and diverse data on the bills can be missing or corrupted. Therefore, imputing missing data is a crucial step in the tool chain.

As decision making, in particular in this sensitive area, should be transparent to both parties, the operational use of black box machine learning algorithms is often seen critically by the stakeholders and is in many cases avoided. As a consequence, rule learning achieving a comparable performance offers the desired advantage of explainability.  

For our case study, we are focusing on \emph{dental bills}. On those bills, the specific type of dental service per row on the bill is unknown but needed for deciding on the amount of refund. Especially differentiating between material costs and other costs is of crucial importance.

In collaboration with the APKV, we have been provided with a training data set consisting of hundreds of thousands of dental bills limiting the application of an ordinary rule learner.

Using only structured information on the bills such as cost, date and simple engineered features, neural network architectures and tree based gradient boosting have been developed achieving an accuracy of 91\%. We apply our methodology on these dental bills in order to obtain competitive results in terms of accuracy that in addition lead to a comprehensible classfication. Note that the derived rules are of great use, even for non-automated classification of such medical bills to achieve more consistency in the decision making.

Table~\ref{tab:bills} shows that the advantages concerning computation time of our approach are not only achieved for some standard benchmark data sets but also for this practical case study. In particular, this application makes the advantages of our approach clear as instead of a computation time of 1-3 days, we are able to find a complete set of rules within a few hours.  

\begin{table}[t]
\centering
\begin{tabular}{@{}|@{\,}l@{\,}|@{\,}l@{\,}|@{\,}r@{\,}|@{\hspace{1mm}}c@{\hspace{1mm}}|@{\hspace{1mm}}r@{\hspace{1mm}}|@{\hspace{1mm}}r@{\hspace{1mm}}|@{\hspace{1mm}}r@{\hspace{1mm}}|@{}}
\hline

\hfil \textbf{data} \hfill & \hfil \textbf{learner} \hfill & \hfil \textbf{time} \hfill & \textbf{speedup} & \textbf{accuracy} & \textbf{precision} & \textbf{recall}
  \\ \hline\hline                                                                                                                      Dental bills & \foil\ &  238.865 s &  & 80,9\% & 50,8\% & 62,5\%\\
\hline
 & \foil\ - mod. & 5.320 s & 44,9 &80,2\%& 50,2\% & 65,7\%\\
\hline
 & \irep\  &  3.588 s & &84,3\% & 76,0\% & 44,8\%\\
\hline
 & \irep\ - mod. & 2.900 s & 1,2 &81,6\% & 62,5\% & 39,0\%\\
\hline
 & \ripper\  & 66.743 s & &86,0\% & 76,9\% & 55,8\% \\
\hline
 & \ripper\ - mod. & 18.140 s & 3,7 &85,9\% & 63,1\% & 52,8\%\\
\hline
\end{tabular}
\smallskip
\caption{Performance of our approach on the case study concerning dental bills.} \label{tab:bills}
\end{table}

Note that precision and recall shown in Table \ref{tab:bills} are arithmetically averaged and appear below value due to poor results corresponding to two very small classes. In future work, we aim to adapt the number of learned rules according to the size of the considered class and take the frequency of each label into account when conducting a multi-class classification.

\section{Alternative Scenarios}
\label{sec:alternative}

Above, we have illustrated and explained our main results, but as already mentioned, one crucial advantage of our approach is the exchangeability of the applied ML tools and rule learners. In this chapter, we summarise our lessons learned considering  the application of various modern clustering and ILP methods. Moreover, we give an outlook of ideas for future work further improving the versatility of our approach, especially in the context of text-based data sets.

\subsection{Alternative Clustering or Representation Learning}

As already mentioned, we decided to apply k-means clustering within our experiments since it has shown to yield the most promising results on the considered benchmark problems. However, applied on other problems, our approach might benefit from different clustering methods. In most of the scenarios, k-means clustering can be easily exchanged by any common clustering method. 

On the other hand, considering text-based data in particular, it makes sense to further exploit the information contained within the text. Therefore, we plan to include language transformer models as, for instance, RoBERTa~\cite{RoBERTa}. These models facilitate straightforward processing of given text data which can also be combined with additional information not contained in the text and yield the features used as input for the subsequently applied clustering method. We believe that text transformers can contribute to opening up new possibilities in the application of rule learners and we aim to further investigate our theory in the course of our collaboration with the APKV, where we will combine the structured information explained in Section~\ref{sec:caseStudy} with additional textual information.

Moreover, Eldbib~\cite{Eldbib:2016} provides an interesting idea for future work in the context of clustering applied for input selection. We plan to adapt his \emph{Data Grouping Method}, which combines clustering with sorting according to a specific \emph{entropy function}, for our purposes and apply it within the choice of negative representatives. Up to now, we applied clustering in order to find as heterogeneous examples as possible and arbitrarily extracted a part of the examples contained in each cluster. We believe that at this point the idea of ordering the negative examples will be advantageous and further improve the rule-learning process. 

\subsection{Alternative Learners}

As noted, our methodology shows that \foil, one of the first ILP methods, has clear advantages over
state-of-the-art rule learning methods like \ripper\ on the considered data sets.
This motivates the question, whether more sophisticated
ILP methods could be employed profitably, for instance, in the context of our case study (see Section~\ref{sec:caseStudy}).
To this purpose, we consider methods that are able to handle noise as faulty classification cannot be excluded in large benchmarks,
cf.~Table~\ref{tab:ilp-comp}, that is, all systems except~\metagol~\cite{metagol}.
However---to our surprise---none of the other methods can, to the best of our knowledge, be suited to the here studied benchmarks.
To exemplify this, we choose the MNIST data set as benchmark. Note that our experiments confirm an achievable accurracy of almost 95\%.

We employ a toy target rule to suitably tune the parameters and guarantee scalability to the MNIST data set. For the images of handwritten digits contained in the MNIST data set it is for instance necessary to learn rules stating whether specific pixels are black or white.
We simplify the task by generating just a few binary vectors of fixed length $N$ with the aim to learn the following target program:
\begin{lstlisting}[style=prolog]
target(V) :- black_1(V), black_N(V).
\end{lstlisting}
Clearly this expresses that the first and the last pixel entry of the target vector should be equal to one (corresponding
to \lstinline[style=prolog]{black_i}), while the remaining entries can be chosen arbitrarily.
The (obvious) definitions of the predicates \lstinline[style=prolog]{black_i} are given as background knowledge of the appropriate
form.

The advantage of this example is that we cannot only observe whether a given method is principally able to learn rules of this form but also change the vector length $N$ in order to find the computational limits of a method because some of the methods discussed below are able to learn simple \emph{black/white rules} on very short vectors but fail as soon as vector length grows beyond $100$. Note that the vectors corresponding to the MNIST data set are of length~$784$.
In the following, we will give an overview of a variety of ILP methods and clarify challenges for their application in our modular methodology.

\paragraph*{Progol.}
\progol\ is a seminal ILP system and has inspired many other approaches, like for example \alephsystem, \metagol\ and many others, cf.~\citep{muggleton1995inverse}.
In our  experiments, we use a freely available \textsf{C} implementation of \progol.%
\footnote{See~\url{https://www.doc.ic.ac.uk/~shm/Software/progol4.4/}.}
While the test program is learnt in a second for the required vector length of $784$, \progol\ fails on the original MNIST data set.
The learned rules overfitted the MNIST data set, essentially always returning the training set as rules.
After applying the same preprocessing steps as Mitra and Baral documented in Section~\ref{sec:eval}, ie.\ using submatrices of size $7 \times 7$, \progol\ returns at least some comprehensible rules (together with hundreds of unpredictable input examples) predicting whether a given digit is equal to zero. The method takes more than 200 hours for a precision of 91.4\% and a recall of 71\%. For comparison, the methods considered in Section~\ref{sec:eval} achieve both precision and recall of more than 98\% and take at most 5 minutes.
We expect that \progol\ mixed search method and essential bottom-up method constitutes the corresponding bottle-neck
which however rendered it unsuitable for application in our modular setup, let alone our case study.

\paragraph*{LIME.}
\lime\ is intended to handle noisy data and constitutes an extension of \foil. We have tried an exisiting \textsf{C}-implementation%
\footnote{See~\url{http://users.cecs.anu.edu.au/~Eric.McCreath/lime/}.}
but this couldn't handle vector lengths ${} > 252$ due to a compile error.
We leave re-implementation of \lime\ in a modern fashion to future work.

\paragraph*{Aleph.}
\alephsystem, is a successor of \progol\ that similarly employs a mixed bottom-up and top-down
strategy.%
\footnote{See~\url{https://www.cs.ox.ac.uk/activities/programinduction/Aleph/aleph.html}.}
Like \progol\ it can handle noise and should thus be poised to handle the MNIST data set. \alephsystem\
is written in Prolog and features an extensive documentation how to employ the leaner effectively.
However, we only managed to get \alephsystem\ to learn the expected \lstinline[style=prolog]{target} predicate for our toy example above on small vector sizes (${} < 15$). Larger vector length immediately lead to overfitting. Thus we aborted the experiment
and did not attempt to try to use \alephsystem\ on (at least parts of) the MNIST data set.

\paragraph*{$\partial$ILP and dNL-ILP.}
\dilp\ has been introduced by Evans and Grefenstette in~\cite{dILP} as a powerful alternative of
classical ILP methods based on \emph{differential learning}.
The core idea is to represent the
problem to synthesis a logic program as a \emph{constraint satisfaction problem}, a common approach in modern
tools, cf~\cite{popper,fastlas}. However, instead of an encoding into SAT or ASP, as would be common
in a standard setting, Evans and Grenfenstette employ a differential implementation of the constraint
satisfaction problem through quantatising. To wit, instead of using propositional logic to
choose a discrete subset of those clauses satisfying the predefined program template, continuous
weights are employed to determine the probability distribution of the clauses. In sum, this
gives rise to the encoding of ILP problems in a neural network architecture.
An extension of this idea has been investigated by Payani et al in~\cite{dNL}, see also~\cite{Payani:2020}
and implemented in the tool~\dnl. These approaches are related to our contribution, as they share the motivation to empower ILP systems
through the use of state-of-the-art methods in machine learning.%
\footnote{We emphasise that wrt. the MNIST data set, Evans et al.~\cite{dILP} learn \emph{comparisons} of handwritten
digits, rather than classification of the digits as in our benchmark. More crucially, this application of~\dilp\
is built upon a pre-trained neural network classifier, so that the final rules learnt do not constitute explainable
rules in our setting.
Further, note that Payani et al.~\citep{dNL} indeed present remarkable results on a IMDB dataset
but they aim to learn the relation \emph{workingUnder(A,B)}, which is a different problem not to be confused with our considered sentiment analysis predicting whether a movie review is positive or negative.}
In our experiments, we employ the respective implementations of \dilp\ and \dnl, available online.%
\footnote{See~\url{https://github.com/ai-systems/DILP-Core} and~\url{https://github.com/apayani/ILP}, respectively.}
Learning the expected \lstinline[style=prolog]{target} predicate for our toy example turned out to be a highly non-trivial task.
One hurdle  is that the background knowledge in both tools has to be given through examples. Hence, one cannot
represent the background knowledge as succinctly for the toy example as above. This we could partly overcome
and manage to learn \lstinline[style=prolog]{target} in tool~\dnl\ for small vector length (${} < 15$).
Beyond that, however, \dnl\ did not produce recognisable rules. In particular the toy target program
couldn't be learnt for the required vector length.
Wrt.\ \dilp\ we also encountered the principal restriction that the tool can only
learn binary Prolog clauses as the memory consumption would be prohibitive to learn (even) ternary clauses,
cf.~\cite{dILP}. However, as attested by our evaluation results, rules learnt for the MNIST data set
constitute of non-binary clauses. Naturally, we could overcome this problem with predicate
invention. Unfortunately, the partial solution to this provided by the existing implementation of \dilp\ is not sufficient here.

\paragraph*{ILASP and FastLAS.}
Partially motivated by Mitra et al.'s use of ASP for their analysis of the MNIST data set,
we investigated the \ilasp~\citep{ilasp} and~\fastlas~\citep{fastlas} systems as ILP systems under the answer set semantics.
Wrt.~\ilasp, we only managed to learn the expected \lstinline[style=prolog]{target} predicate for our toy example above on small vector sizes (${} \leqslant 5$), while we could not obtain positive results with~\fastlas. In hindsight this is not too surprising as the bottom-up approach of both systems yields an
enormous search space even for our toy example. This may also explain, why in~\cite{mitra18} a pre-processing of the MNIST
data set is performed, yielding, as described above, a $7 \times 7$ matrix representation of the original data. As mentioned
in~\cite{mitra18}, the employed \xhail\ system~\cite{Ray09} was not able to handle the data set in its original form.

\paragraph*{Popper.}
The \popper\ system~\cite{popper} combines ideas from counter-example guided program synthesis with the
use of ASP as constraint solver. This yields a very powerful ILP system that is well-equipped
to learn even higher-order programs (see also~\cite{Purgal:2021}).
However, wrt. the MNIST data set or even the toy example considered, the machinery is not well-suited. Indeed,
again we were only able to learn \lstinline[style=prolog]{target} predicate for our toy example above on small vector sizes (${} < 100$).
This is in accordance with the theoretical investigations by Cropper et al.~\cite[Proposition~1]{popper} detailing that the hypothesis space grows exponentially, that is, in our case exponentially in the size of the target vector. Infeasibilty
of the approach for a vector size of $784$ is thus hardly surprising.

\section{Related Work}
\label{Related}

The key contribution of our work is a modular approach which makes it possible to apply existing rule learning methods in reasonable time on very large data sets. 
In particular, we have conducted detailed experiments on the MNIST, Fashion-MNIST and IMDB data sets as
well on a particular case study stemming from industrial collaborations.
There are only few approaches \citep{nguyen, burkhardt, granmo2019convolutional} which overlap with our considerations to a certain extent. They all have in common that they also consider the MNIST and  Fashion-MNIST data set as well as similar ones with the main goal of interpretability of the results. 
Regarding sentiment analysis with the example of the IMDB data set, to the best of our knowledge there are no approaches similar to ours out there.

Nguyen et al.~\cite{nguyen} learn features from given training data by least generalisation. Their results look similar to our rules depicted in Figure \ref{fig:mnist_rules} with the difference that they rather learn a kind of prototype of each class than deterministic rules. For classification, they basically compare the input image with each learned prototype and predict the one where the sum of differences between pixel values in each location is minimal. On the MNIST data set they obtain a precision of 66\% and a recall of 80\%, where we achieve more than 94\% in both metrics. On the Fashion-MNIST data set they obtain 61\% and 63 \% in precision and recall compared with over 84\% achieved by our approach. They do not go into further detail about the time consumption of their experiments.

Burkhardt et al.~\cite{burkhardt} deal with the extraction of rules from binary neural networks. On the one hand they seek to find rules explaining the decisions of a learned neural network and evaluate their results with a corresponding \emph{similiarity measure}, but on the other hand they also use the original labels of the data sets to learn rules. In the latter approach they achieve accuracies between 83\% and 94\% on the MNIST data set compared with 88-94\% from us and on the Fashion-MNIST data set they obtain between 82\% and 88\%, where we reach 80-85\%. They do not mention the consumed time in further detail but we assume that their approach takes significantly longer than the 300-500 seconds needed for our approach since a neural network has to be trained first.

Granmo et al.~\cite{granmo2019convolutional} introduce the \emph{Convolutional Tsetlin Machine} as an interpretable alternative to CNNs. Basically, they learn several patterns which are used like filters in a CNN but which are easier to interpret, although they seem to be not as  self-explaining as our learned rules. So, their approach is probably not as easily interpretable as ours but compares favorably with simple CNNs achieving an accuracy of 99\% on MNIST and 91\% on Fashion-MNIST. The execution time of their approach is strongly dependent of the number of allowed clauses per class. They learn the patterns epoch-per-epoch and report execution times of 39 seconds per epoch on MNIST and 52 seconds on Fashion-MNIST using a NVIDIA DGX-2 server. Usually, they learn for about 200 epochs, which indicates that their approach in general takes longer than ours to yield results, speaking of 8 thousand to 10 thousand seconds which corresponds to a speedup of factor 20-30 using our approach.  

\section{Conclusion \& Future Work} \label{sec:future}

In this paper, we present a modular methodology that combines state-of-the-art methods in non-explainable machine learning with traditional methods in rule learning to provide efficient and scalable algorithms for the classification of vast data sets, while still achieving explainability through the provision of
comprehensible rules. We show the viability and competitiveness of the introduced methodology
in the context of large scale data sets and an industrial case study.

Future work will be concerned with the extension of the representation learning and
input selection. Wrt. the former, we want to incorporate the use of text transformers, as already indicated above, to extend the applicability of our results on the case study. Wrt. the latter, we want to investigate
more sophisticated clustering approaches, for example taking the approach in~\cite{Eldbib:2016} into account.

\backmatter


\section*{Declarations}

\paragraph*{Funding}

Partial financial support was received by the doctoral scholarship from the promotion of young researchers at the University of Innsbruck (\emph{Doktoratsstipendium aus der Nachwuchsförderung der Universität Innsbruck}) and by the \emph{metafinanz Informationssysteme GmbH} on behalf of the \emph{Allianz Private Krankenversicherung} in the course of the project \emph{Modelle zur Ergänzung fehlender Daten in der Allianz Private Krankenversicherung} at the University of Innsbruck.

\paragraph*{Conflict of interest}

The authors have no competing interests to declare that are relevant to the content of this article.

\paragraph*{Ethics approval}
Not applicable.

\paragraph*{Consent to participate}
Not applicable.

\paragraph*{Consent for publication}
Not applicable.

\paragraph*{Availability of data and materials}

The data sets \begin{inparaenum}[(i)]
\item \emph{MNIST digits}~\citep{lecun2010mnist};
\item \emph{Fashion-MNIST}~\citep{fmnist}; and
\item \emph{IMDB}~\citep{imdb},
\end{inparaenum}
are freely available, as indicated in the references. The data set
stemming from the industrial collaboration cannot be made available, due to pending
non-disclosure agreements.

\paragraph*{Code availability}

The code and the experimental data are freely availabe at the indicated github page~\url{https://github.com/AlbertN7/ModularRuleLearning}.

\paragraph*{Authors' contributions}

All authors contributed to the study conception and design. Material preparation, data collection and analysis were performed by Albert Nössig. The first draft of the manuscript was written by Albert Nössig and all authors commented on previous versions of the manuscript. All authors read and approved the final manuscript.

\renewcommand{\bibfont}{\small}
\setlength{\bibsep}{3pt}

\newpage
\appendix

\section{Rule Learning Algorithms}
\label{Techniques}

Above, we have introduced a modular approach combining rule learning methods with additional (unexplainable) machine learning tools and illustrated the achieved advantages, in particular the gained time savings. 
In the following, we revise the main ideas behind the rule learners successfully applied on the data sets of interest, ie. the \foil\ algorithm as well as the \ripper\ algorithm. In addition to self-certainty, this serves to compare the algorithms in more detail as well as to give some insights in our implementation of the \foil\ algorithm.

\paragraph*{FOIL~\citep{DBLP:journals/ml/Quinlan90}.}

\emph{First Order Inductive Learner} (\foil) is one of the first ILP systems and despite its simplicity, it performs quite remarkably on our experiments yielding equivalent results to state-of-the-art rule learning methods. We believe that its simplicity in particular makes it applicable to many problems, where modern methods fail as shown in Section~\ref{sec:alternative}. The basic procedure, given a set of positive examples $\mathcal{P}$ and a set of negative examples $\mathcal{N}$, is outlined in the upper half of Figure~\ref{fig:foil_ripper}.
Note that the set of possible literals depends on the specific input data and consists of conditions of the form $a = v$ for all attributes $a$ that are not yet included in the considered \emph{NewRule} and all values $v$ occurring in the input data as further explained in Section~\ref{sec:eval}. Moreover, within the \foil\ algorithm a specific \emph{gain function} is applied in order to find the best possible literal. Basically, this means that a literal should cover as much positive examples as possible and at the same time as few negative ones as possible. The performance of the \foil\ algorithm is in particular dependent on the chosen gain function \citep{Jimnez2011OnIF}.

For the experiments conducted in Section~\ref{sec:eval}, we implemented the \foil\ algorithm on our own because we wanted a running Python implementation
on the one hand and on the other hand our implementation simplifies the application of the algorithm to large, array-like data sets as in particular in the considered case study.%
\footnote{The corresponding code is available at~\url{https://github.com/AlbertN7/ModularRuleLearning}.} 

\paragraph*{RIPPER~\citep{COHEN1995115} and IREP~\citep{Frnkranz1994IncrementalRE}.}
\emph{Repeated Incremental Pruning to Produce Error Reduction} (\ripper) is a rule learning method which is able to handle large noisy data sets effectively. Similar as its predecessor \emph{Incremental Reduced Error Pruning} (\irep\ by~\cite{Frnkranz1994IncrementalRE}), it combines the separate-and-conquer technique applied in \foil\ with the  reduced error pruning strategy \citep{Brunk1991AnIO}. The pseudocode for the algorithm is outlined in the lower half of Figure~\ref{fig:foil_ripper}, where $\mathcal{P}$ and $\mathcal{N}$ again denote the positive and negative examples, respectively.

Note that the \emph{description length} is a measure of total complexity (in bits) preventing overfitting. It consists of the complexity of the model plus that of the examples which are not captured. As the learned set of rules grows, the complexity of the model increases and the number of uncovered examples decreases. In sum one can say that description length helps to balance between minimisation of classification error and minimisation of model complexity.
The \emph{growing set} typically consists of two third of the input data, whereas the \emph{pruning set} contains the remaining third. As the names already suggest, the growing set is used during the growing phase, where conditions are iteratively added to a rule similar as in the \foil\ algorithm until no negative examples are covered. On the other hand, the pruning set is used during the pruning phase, where the learned conditions are deleted in reversed order to find the pruned rule which maximizes $\frac{p - n}{p + n}$. Here, $p$ and $n$ denote the number of positive and negative pruning examples that are covered by the pruned rule. 

One crucial difference of the \ripper\ algorithm opposed to the \irep\ algorithm is the additional optimisation phase, which is applied for $k$ iterations after the generation of a rule set. Typically $k$ is equal to 2. During the optimisation each rule is considered in the order in which they have been learned and two alternative rules are constructed for each original rule. The \emph{replacement rule} is learned by growing and pruning from scratch, whereas the \emph{revision rule} is grown by greedily adding conditions to the considered original rule rather than the empty rule. Afterwards, a description length heuristic is used to choose the better one.
In our experiments we made use of the \emph{wittgenstein} package implemented in Python.

\begin{figure}[h]
\begin{lstlisting}
@(\textbf{FOIL(}$\mathcal{P}, \mathcal{N}$\textbf{)})@
LearnedRules <- {}
while (@($\mathcal{P}$)@ not empty):
	//@(\textbf{Learn a new rule})@
	NewRule <- Rule without any conditions
	CoveredNegs <- @($\mathcal{N}$)@
	while (CoveredNegs not empty):
		//@(\textbf{Add a literal to NewRule})@
		PossibleLiterals <- Set of all possible literals 
		BestLiteral <- @($\argmax_{l~\in~\text{PossibleLiterals}}$ \textbf{Gain(}$l$\textbf{)})@
		NewRule.append(BestLiteral)
		CoveredNegs <- Subset of CoveredNegs satisfying NewRule
	LearnedRules.append(NewRule)
	@($\mathcal{P}$)@ <- Subset of @($\mathcal{P}$)@ not covered by LearnedRules
	
return LearnedRules
\end{lstlisting}
\begin{lstlisting}
@(\textbf{GenerateRuleSet(}$\mathcal{P}, \mathcal{N}$\textbf{)})@
LearnedRules <- {}
DL <- DescriptionLength(LearnedRules,@($\mathcal{P}, \mathcal{N}$)@)
while (@($\mathcal{P}$)@ not empty):
	//@(\textbf{Grow and prune a new rule})@
	split (@($\mathcal{P}, \mathcal{N}$)@) into (@($\mathcal{P}_{\text{grow}}, \mathcal{N}_{\text{grow}}$)@) and (@($\mathcal{P}_{\text{prune}}, \mathcal{N}_{\text{prune}}$)@)
	NewRule <- GrowRule(@($\mathcal{P}_{\text{grow}}, \mathcal{N}_{\text{grow}}$)@)
	NewRule <- PruneRule(NewRule,@($\mathcal{P}_{\text{prune}}, \mathcal{N}_{\text{prune}}$)@)
	LearnedRules.append(NewRule)
	if (DescriptionLength(LearnedRules,@($\mathcal{P}, \mathcal{N}$)@) > DL + 64):
		 //@(\textbf{Prune the whole rule set and exit})@
		 for each rule @($\mathcal{R}$)@ in LearnedRules in reversed order:
		 	if (DescriptionLength(LearnedRules\{@($\mathcal{R}$)@},@($\mathcal{P}, \mathcal{N}$)@) < DL):
		 		LearnedRules.delete(@($\mathcal{R}$)@)
		 		DL <- DescriptionLength(LearnedRules,@($\mathcal{P}, \mathcal{N}$)@)
		return LearnedRules
	DL <- DescriptionLength(LearnedRules,@($\mathcal{P}, \mathcal{N}$)@)
	@($\mathcal{P}$)@ <- Subset of @($\mathcal{P}$)@ not covered by LearnedRules
	@($\mathcal{N}$)@ <- Subset of @($\mathcal{N}$)@ not covered by LearnedRules

return LearnedRules
		
@(\textbf{OptimizeRuleSet(}LearnedRules,$\mathcal{P}, \mathcal{N}$\textbf{)})@ 
for each rule @($\mathcal{R}$)@ in LearnedRules:
	LearnedRules.delete(@($\mathcal{R}$)@)
	@($\mathcal{P_{\text{unc}}}$)@ <- Subset of @($\mathcal{P}$)@ not covered by LearnedRules
	@($\mathcal{N_{\text{unc}}}$)@ <- Subset of @($\mathcal{N}$)@ not covered by LearnedRules
	split (@($\mathcal{P_{\text{unc}}}, \mathcal{N_{\text{unc}}}$)@) into (@($\mathcal{P}_{\text{grow}}, \mathcal{N}_{\text{grow}}$)@) and (@($\mathcal{P}_{\text{prune}}, \mathcal{N}_{\text{prune}}$)@)
	ReplRule <- GrowRule(@($\mathcal{P}_{\text{grow}}, \mathcal{N}_{\text{grow}}$)@)
	ReplRule <- PruneRule(ReplRule,@($\mathcal{P}_{\text{prune}}, \mathcal{N}_{\text{prune}}$)@)
	RevRule <- GrowRule(@($\mathcal{P}_{\text{grow}}, \mathcal{N}_{\text{grow}},\mathcal{R}$)@)
	RevRule <- PruneRule(RevRule,@($\mathcal{P}_{\text{prune}}, \mathcal{N}_{\text{prune}}$)@)
	OptimizedRule <- better of ReplRule and RevRule
	LearnedRules.append(OptimizedRule)
	
return LearnedRules
		 	
@(\textbf{RIPPER(} $\mathcal{P}, \mathcal{N}, k$\textbf{)})@
LearnedRules <- @(\textbf{GenerateRuleSet(}$\mathcal{P}, \mathcal{N}$\textbf{)})@
repeat k times:
	LearnedRules <- @(\textbf{OptimizeRuleSet(}LearnedRules,$\mathcal{P}, \mathcal{N}$\textbf{)})@ 
	
return LearnedRules
\end{lstlisting}
\caption{Pseudocode Describing the \foil\ and the \ripper\ Algorithm.} \label{fig:foil_ripper}
\end{figure}

\end{document}